\documentclass[conference]{IEEEtran}
\IEEEoverridecommandlockouts
\usepackage{cite}
\usepackage{amsmath,amssymb,amsfonts}
\usepackage{algorithmic}
\usepackage{graphicx}
\usepackage{textcomp}
\usepackage{xcolor}
\usepackage{booktabs}
\def\BibTeX{{\rm B\kern-.05em{\sc i\kern-.025em b}\kern-.08em
    T\kern-.1667em\lower.7ex\hbox{E}\kern-.125emX}}
\begin{document}

\title{ResNet: Enabling Deep Convolutional Neural Networks through Residual Learning\\}

\author{\IEEEauthorblockN{Xingyu Liu}
\IEEEauthorblockA{\textit{School of Computing} \\
\textit{Singapore Polytechnic}\\
Singapore, Singapore \\
https://orcid.org/0009-0004-6556-5229}
\and
\IEEEauthorblockN{Kun Ming Goh}
\IEEEauthorblockA{\textit{School of Computing} \\
\textit{Singapore Polytechnic}\\
Singapore, Singapore \\
https://orcid.org/0009-0008-7666-781X}

}

\maketitle

\begin{abstract}
Convolutional Neural Networks (CNNs) have revolutionised computer vision, but training very deep networks has been challenging due to the vanishing gradient problem. This paper explores Residual Networks (ResNet), introduced by He \emph{et~al.} (2015), which overcome this limitation by using skip connections. ResNet enables the training of networks with hundreds of layers by allowing gradients to flow directly through shortcut connections that bypass intermediate layers. In our implementation on the CIFAR-10 dataset, ResNet-18 achieves 89.9\% accuracy compared to 84.1\% for a traditional deep CNN of similar depth, while also converging faster and training more stably.
\end{abstract}


\section{Introduction}
Deep Convolutional Neural Networks (CNNs) have become the foundation of modern computer vision, powering applications from image classification to object detection. Traditional CNN architectures consist of sequential layers, including convolutions, pooling, and fully connected layers. However, when training very deep networks, the problem of vanishing gradients becomes a major obstacle.

This issue arises when gradients shrink exponentially as they are backpropagated through many layers, preventing earlier layers from learning effectively. Surprisingly, the addition of more layers can decrease the performance rather than improve it, a phenomenon known as the \emph{degradation problem}. Before ResNet, most CNNs could not effectively exceed 20--30 layers; architectures such as VGG-16 and VGG-19 approached this limit using conventional training methods. The introduction of Residual Networks (ResNet) in 2015 marked a turning point, making it possible to train networks with 50, 101, or even 152 layers.

The key innovation in ResNet is the use of residual (or skip) connections. Rather than learning a direct mapping from input to output, each block learns a residual mapping, which is the difference between the desired output and the input. This design facilitates gradient flow during training and allows for the construction of deeper networks without degradation in performance.

In this paper, we examine ResNet's architecture, implementation, and performance benefits. We show that residual connections not only enable deeper networks, but also result in more stable training and better accuracy on image classification tasks like CIFAR-10.

\section{Related Work}
\subsection{Evolution of Deep CNN Architectures}
Several key developments have shaped the evolution of deep CNNs. LeNet-5 \cite{lecun1998gradient} introduced the basic CNN structure with convolutional, pooling, and fully connected layers. AlexNet \cite{krizhevsky2012imagenet} demonstrated the power of deeper networks combined with GPU acceleration, achieving breakthrough results on ImageNet with an eight-layer model.

VGG networks \cite{simonyan2014very} further pushed depth with 16--19 layers, confirming that deeper architectures tend to perform better while highlighting challenges such as vanishing gradients and high computational cost. GoogLeNet/Inception \cite{szegedy2014going} introduced inception modules with multiple parallel convolution paths, allowing for a deep yet efficient architecture.

\subsection{Addressing the Vanishing Gradient Problem}
Batch normalisation \cite{ioffe2015batch} stabilises training by normalising the input of the layers, enabling faster convergence and higher learning rates. Highway Networks \cite{srivastava2015training} introduced gating mechanisms that allow information to pass through layers unchanged; this work directly inspired ResNet, but incurred additional architectural overhead due to learnt gates.

\subsection{Residual Learning Framework}
ResNet \cite{he2016deep} introduced residual learning: instead of learning a function $H(x)$, each layer learns a residual function $F(x)=H(x)-x$, resulting in the output $F(x)+x$. This makes it easier for layers to approximate identity mappings when necessary, allowing gradients to flow directly through skip connections. The original paper demonstrated networks up to 152 layers with state-of-the-art results on ImageNet and CIFAR-10.

\subsection{Extensions and Improvements}
ResNet inspired several extensions. ResNeXt \cite{xie2016aggregated} combines residual learning with grouped convolutions to achieve accuracy gains at a similar cost. DenseNet \cite{huang2017densely} connects each layer to every other layer in a feed-forward manner. Wide ResNet \cite{zagoruyko2016wide} increases the width of the network rather than the depth to achieve comparable or better performance with fewer layers.

\section{Dataset and Methodology}
\subsection{Dataset Description}
We use CIFAR-10, a widely used benchmark with 60{,}000 $32{\times}32$ colour images across 10 classes (50{,}000 train / 10{,}000 test). CIFAR-10 provides sufficient complexity to demonstrate the benefits of deeper networks while remaining computationally tractable for repeated experiments.

\subsection{Data Preprocessing}
We apply standard preprocessing and augmentation: (i) pixel values scaled to $[-1,1]$ using $(x-0.5)/0.5$; (ii) labels one-hot encoded; (iii) random horizontal flips and brightness jitter (max\_delta $=0.1$); and (iv) an efficient \texttt{tf.data} input pipeline with shuffling, parallel mapping, batching, and prefetching.

\subsection{Network Architectures}
\subsubsection{Baseline CNN}
Four convolutional blocks with filters $32{\rightarrow}64{\rightarrow}128{\rightarrow}256$; each block uses Conv2D~$\rightarrow$~BatchNorm~$\rightarrow$~ReLU~$\rightarrow$~MaxPooling. A global average pooling layer feeds a dense classifier; dropout ($p{=}0.5$) regularises the head. Total parameters: $\sim$392k.

\subsubsection{Mini-ResNet}
An efficient ResNet-style model: initial 32-filter convolution; four stages with filters $[32,64,128,256]$; each stage has two residual blocks, the first using stride~2 and a projection shortcut. Global average pooling, dropout, and a softmax classifier. Total parameters: $\sim$2.8M.

\subsubsection{ResNet-18 (Custom)}
Adapted to CIFAR-10: initial 64-filter convolution; four stages with filters $[64,128,256,512]$, two residual blocks per stage; the first block in each stage uses projection shortcuts for downsampling. Batch Normalisation and ReLU throughout; global average pooling, dropout, and a softmax classifier. Total parameters: $\sim$11M.

\subsection{Training Configuration}
All models are trained with Adam (initial learning rate $10^{-3}$); \texttt{ReduceLROnPlateau} (factor $=0.2$, patience $=3$); batch size 64; up to 30 epochs with early stopping (patience $=7$) on validation loss; categorical cross-entropy; dropout ($p{=}0.5$) before the final dense layer; and \texttt{ModelCheckpoint} for best weights.

\subsection{Evaluation Metrics}
We report Top-1 accuracy, training stability (accuracy/loss curves), computational efficiency (time per epoch and memory), and confusion matrix / classification report for class-wise analysis.

\section{Results and Discussion}
\subsection{Classification Performance}
Table~\ref{tab:mainresults} summarises the results. The ResNet-18 model achieves the highest accuracy at 89.9\%, improving by 5.8 percentage points over the baseline CNN. Despite having more parameters, ResNet-based models converge more quickly and stably, confirming the efficacy of skip connections for training deeper networks.

\begin{table}[!t]
\caption{Model comparison on CIFAR-10.}
\label{tab:mainresults}
\centering
\begin{tabular}{lccc}
\toprule
\textbf{Model} & \textbf{Test Acc.} & \textbf{Train Time (min)} & \textbf{Params} \\\midrule
Baseline CNN & 84.1\% & 2 & 392k \\
Mini-ResNet & 87.0\% & 7 & 2.8M \\
ResNet-18 (With Skip) & 89.9\% & 24 & 11.2M \\
ResNet-18 (No Skip) & 86.0\% & 210 & 11.0M \\
\bottomrule
\end{tabular}
\end{table}

\begin{figure}[htbp]
    \centering
    \includegraphics[width=0.48\textwidth]{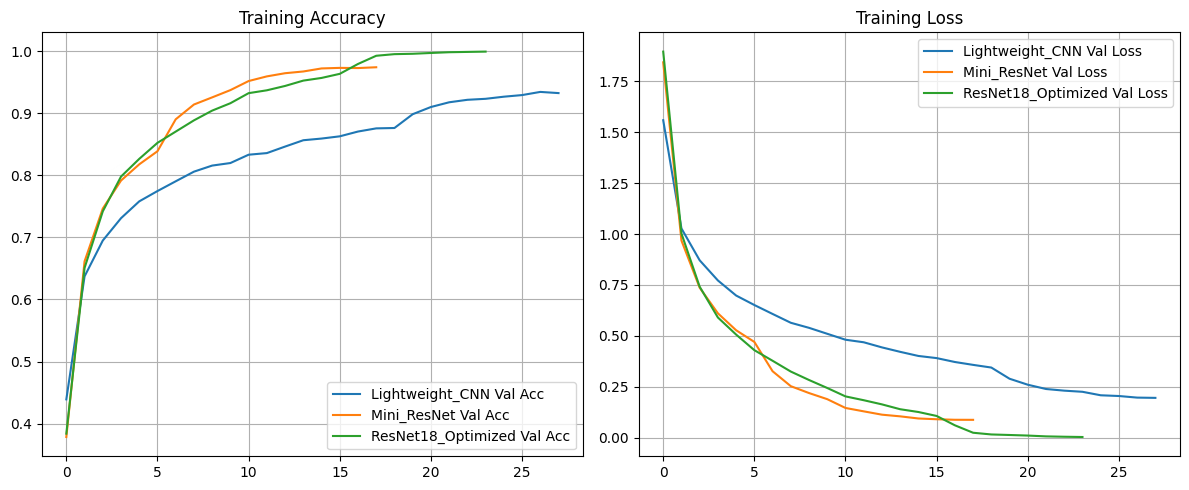}
    \caption{Training accuracy and loss trends for all models. The ResNet-18 model (green) achieves faster convergence and lower training loss, indicating improved gradient flow and stable optimisation compared to the baseline CNN.}
    \label{fig:training_curve}
\end{figure}

\begin{figure}[htbp]
    \centering
    \includegraphics[width=0.48\textwidth]{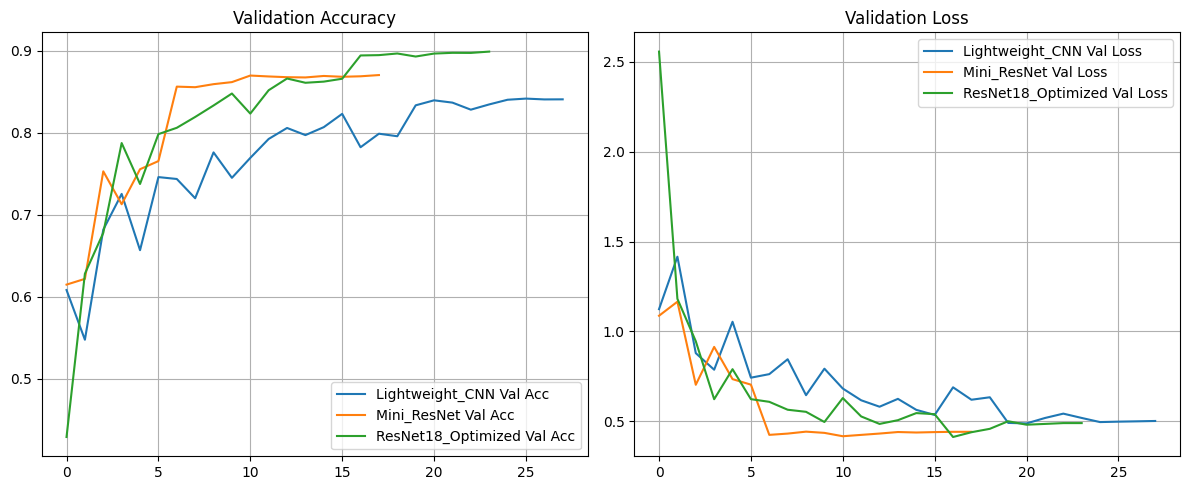}
    \caption{Validation accuracy and loss trends for all models. The ResNet-18 (green) converges faster and achieves the highest validation accuracy with the lowest validation loss, demonstrating the benefit of residual connections.}
    \label{fig:validation_curve}
\end{figure}

\subsection{Training Dynamics}
The baseline CNN exhibits slower convergence and minor overfitting; improvements plateau early, and the training is sensitive to learning rate and initialisation. In contrast, Mini-ResNet and ResNet-18 display smooth, consistent loss reduction and steadily improving validation accuracy, with reduced sensitivity to hyperparameters.

\subsection{Gradient Flow Analysis}

\begin{figure}[htbp]
    \centering
    \includegraphics[width=0.48\textwidth]{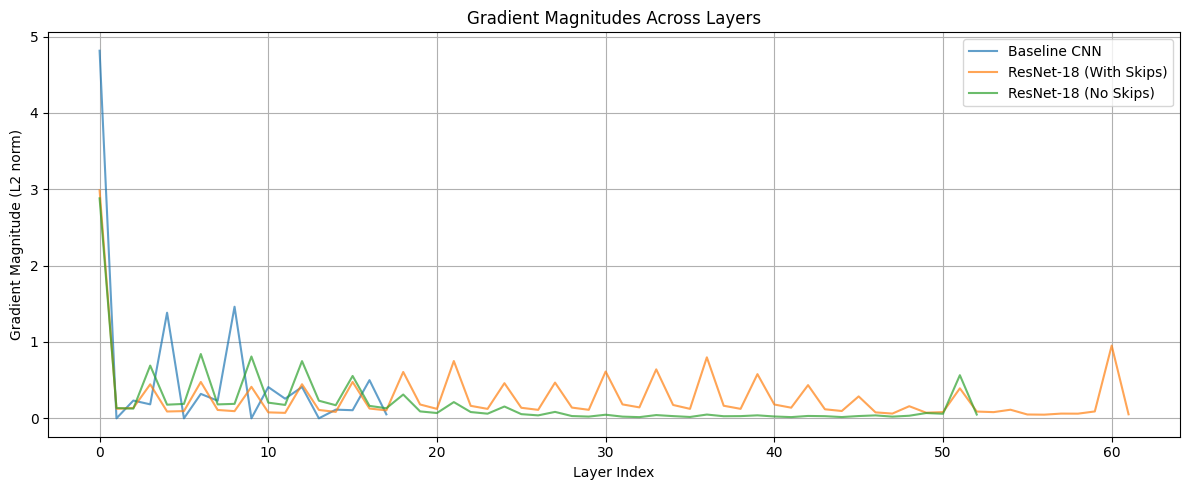}
    \caption{Gradient magnitude distribution across layers for different models. The ResNet-18 with skip connections maintains stable gradient flow throughout the network, while the baseline CNN and non-residual variant experience severe vanishing gradients in earlier layers.}
    \label{fig:gradient_magnitudes}
\end{figure}

We analyse gradient magnitudes (L2 norm) across layers after training. The CNN of the baseline shows a sharp drop in the early layers, indicative of vanishing gradients. ResNet-18 maintains more uniform gradient magnitudes; skip connections permit strong gradients to reach early layers. Removing residual connections collapses gradient flow, confirming their need for effective optimisation.

\subsection{Ablation Study}

\begin{figure}[htbp]
    \centering
    \includegraphics[width=0.48\textwidth]{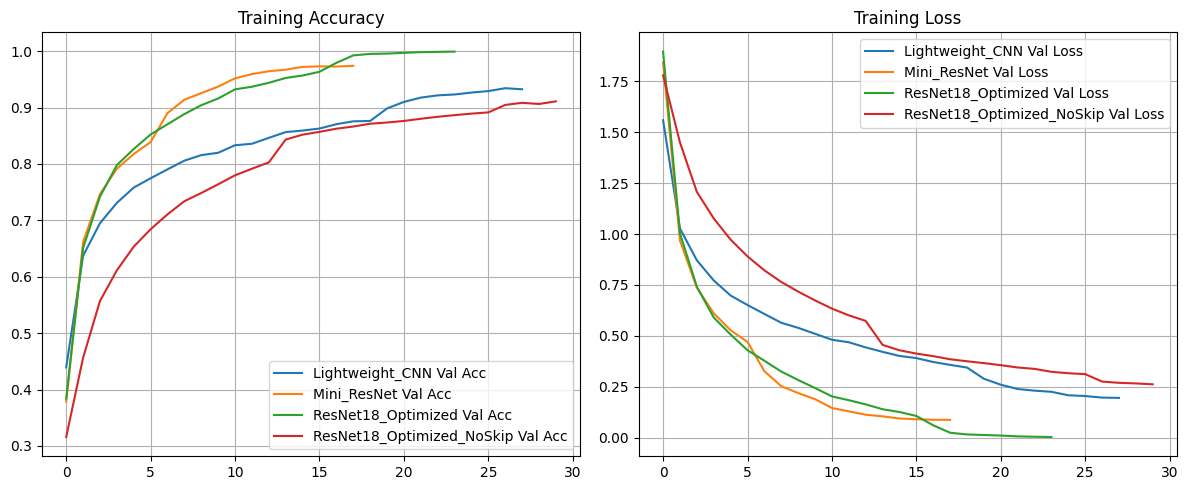}
    \caption{Training accuracy and loss across all models. The ResNet-18 with skip connections converges faster and maintains lower loss throughout training, while the no-skip variant suffers from slower convergence and higher residual error.}
    \label{fig:training_curve_all}
\end{figure}

\begin{figure}[htbp]
    \centering
    \includegraphics[width=0.48\textwidth]{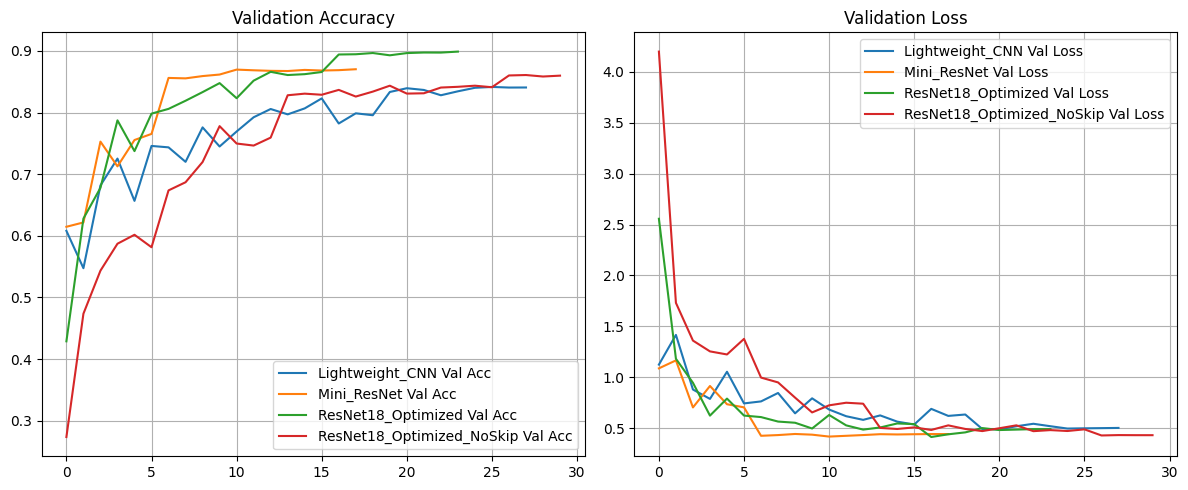}
    \caption{Validation accuracy and loss across all models, including ResNet-18 variants. The ResNet-18 with skip connections achieves the highest validation accuracy and lowest loss, confirming the effectiveness of residual learning compared to the no-skip variant and other baselines.}
    \label{fig:val_curve_all}
\end{figure}

Removing skip connections from ResNet-18 while keeping depth and structure constant degrades accuracy and gradient flow. Thus, residual connections are not optional enhancements but critical components for training deep networks.

\subsection{Computational Considerations}
Although ResNet models have higher parameter counts, they are computationally efficient in practice: (i) memory overhead of skip connections is modest; (ii) faster convergence reduces the number of required epochs; and (iii) inference latency remains similar across models for $32{\times}32$ inputs.

\subsection{Comparison with Related Work}
Our findings align with the original ResNet paper: the $\sim$5--6\% improvement in CIFAR-10, improved convergence stability, and preserved gradient magnitudes are consistent with the established literature.

\section{Conclusion}
Residual connections substantially improve the trainability and performance of deep CNNs. Through systematic evaluation on CIFAR-10, we confirm that residual learning enables superior accuracy, stable optimisation, and practical depth scaling. ResNet remains a cornerstone of modern computer vision and continues to inspire architectural innovation.

\bibliographystyle{IEEEtran}

\end{document}